\documentclass{article}

\usepackage{arxiv}

\usepackage[utf8]{inputenc} 
\usepackage[T1]{fontenc}    
\usepackage{hyperref}       
\usepackage{url}            
\usepackage{booktabs}       
\usepackage{amsfonts}       
\usepackage{nicefrac}       
\usepackage{microtype}      
\usepackage{graphicx}
\usepackage[numbers]{natbib}
\usepackage{doi}

\usepackage[edges]{forest}
\usetikzlibrary{shadows,arrows.meta}
\tikzset{
  parent/.style={align=center,text width=8cm,fill=gray!50,rounded corners=2pt},
  child/.style={align=center,text width=2.5cm,fill=gray!20,rounded corners=6pt},
  grandchild/.style={fill=white,text width=2.0cm}
}

\title{A Survey of Performance Optimization in Neural Network-Based Video Analytics Systems}

\author{Nada Ibrahim\\
        Department of Computer Science\\
        New Mexico State University\\
        NM, USA\\
        \texttt{nada@nmsu.edu}
        
        \And
        Preeti Maurya\\
        Department of Computer Science\\
        New Mexico State University\\
        NM, USA\\
        \texttt{preema@nmsu.edu}

        \And
        \href{https://orcid.org/0000-0003-3422-2755}{\includegraphics[scale=0.06]{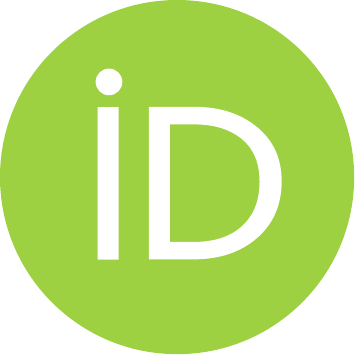}\hspace{1mm}Omid Jafari}\\
        Department of Computer Science\\
        New Mexico State University\\
        NM, USA\\
        \texttt{ojafari@nmsu.edu}
        
        \And
        \href{https://orcid.org/0000-0001-6284-9251}{\includegraphics[scale=0.06]{orcid.pdf}\hspace{1mm}Parth Nagarkar}\\
        Department of Computer Science\\
        New Mexico State University\\
        NM, USA\\
        \texttt{nagarkar@nmsu.edu}        
        }
        
\date{}

\begin{document}
\maketitle

\begin{abstract}
Video analytics systems perform automatic events, movements, and actions recognition in a video and make it possible to execute queries on the video. As a result of a large number of video data that need to be processed, optimizing the performance of video analytics systems has become an important research topic. Neural networks are the state-of-the-art for performing video analytics tasks such as video annotation and object detection. Prior survey papers consider application-specific video analytics techniques that improve accuracy of the results; however, in this survey paper, we provide a review of the techniques that focus on optimizing the performance of Neural Network-Based Video Analytics Systems. 
\end{abstract}


\section{Introduction}
\noindent Over the last decade, the amount of video data that need to be processed has increased exponentially due to the frequent use of cameras everywhere \cite{lu2016optasia, xu2019vstore, moll2020exsample, poms2018scanner}. A wide range of propitious applications such as intelligent transportation systems, security systems, augmented/virtual reality, and advanced driving assistance systems rely on video analytics. Video analytics deals with automatic recognition of temporal and spatial events in videos. For instance, video analytics recognizes and detects humans and objects automatically in a video stream. Moreover, it recognizes human movements, actions, describes and captions videos, and classifies activities and objects. Recently, machine learning and computer vision have been used in multiple domains and they have revolutionized the field of video analytics. Applications use video analytics to take immediate actions based on their decision without any human interactions. Besides, several database management systems that work with data and query processing on videos have more advanced features \cite{daum2020tasm} to support video analytics. 

\noindent Although video analytics is one of the popular research areas now, it is still challenging to get the required accuracy and performance with algorithms available up to date. Query processing in video streams needs to be fast, precise, and scalable. It is important for the algorithms used in video analytics to be efficient enough for huge amount of data. For instance, while many applications require to produce query results in real-time, others can permit a lag of even many minutes to process queries. This allows for temporarily reallocating some resources from the lag-tolerant queries during the interim shortage of resources. Such shortage happens due to a burst of new video queries or “spikes” in resource usage of existing queries (e.g. due to an increase in the number of cars to track on the road) \cite{zhang2017live}. 

\subsection{Motivation}
\subsubsection{Motivation for using Video Analytics}
\noindent Motion detection, object tracking, and scene analysis are all considered as video analytics. Motion detection is to compare the current image with the static background of the scene. Object tracking finds a specific object in the current frame that relates to the object in the next frame. Scene analysis is to recognize actions in the scene. Video analytics helps in understanding the situation in a video and predict the next steps by tracking objects in the video. The detected behaviour of the object in the video helps users to take actions accordingly. Another example of video analytics applications is autonomous vehicles. Self-driving cars have to detect objects in real-time to avoid accidents and collisions. Therefore, optimizations in video analytics are required to reach higher accuracy and lower latency. 

\noindent Moreover, researchers use video analytics techniques to recognize the most common occurring road factors that influence the interaction between autonomous vehicles and other traffic participants \cite{madigan2019understanding}. Video analytics is also crucial in improving safety and security. For instance, behavior detection by determining a person’s posture is used in safety and security applications to detect if a person has fallen, crouched down, or jumping over barriers \cite{okita2020ai}.

\subsubsection{Motivation of our work (difference with other surveys)}
\noindent The current video analytics surveys are on the spatio-temporal and content-based viewpoints \cite{shih2017survey}. Some papers such as \cite{wang2003video, zhang2018physics, kong2018human} focus on video analytics technologies of human behavior and actions. Other works such as \cite{shih2017survey, cuevas2020techniques} focus on video analytics for sports. \cite{olatunji2019video} is a survey that focuses on video analytics and techniques for surveillance camera data and categorizes video analytics subdomains as behavior analysis, moving object classification, video summarization, object detection, object tracking, and congestion analysis. \cite{zhang2019edge} is another survey that focuses on reviewing the video analytics algorithms used in public safety. \cite{zhang2020machine} is a survey that focuses on techniques and methods for optimizing video coding, which is a video content representation format used for storing and transmitting video data. Moreover, there is a survey on human group activity recognition by analyzing person actions from video sequences using machine learning techniques \cite{kulkarni2020survey}. In \cite{premkumar2021video}, the modern deep learning based video analytics approaches are compared with the standard Computer Vision based approaches for Internet of Things (IoT) devices.
\noindent As mentioned earlier, the previous surveys have reviewed application-specific methods; however, they do not include general-purpose video analytics techniques that are focused on optimizing the end-to-end performance of video analysis. In this survey paper, we focus only on Performance Optimization in Video Analytics Systems and we review how these systems work and the optimization strategy that they use. Other works such as \cite{suprem2020odin, ran2018deepdecision, bastani2020vaas, poddar2020visor} focus on improving the accuracy and privacy; hence, they are beyond the scope of this survey paper and we do not include them in our review.

\subsection{Contributions}
\label{sec:contributions}
\noindent In this paper, we present an in-depth review of the recent advances in Optimization-based Video Analytics techniques. Our contributions are listed as following:
\begin{itemize}
	\item We perform an in-depth review over Optimization-based Video Analytics techniques. Our review consists of the definitions, the workflow, and the ideas which are proposed in each technique to improve the end-to-end performance.
	
	\item We categorize the reviewed techniques based on the optimization strategy that they utilize.
\end{itemize}

\subsection{Paper Organization}
\label{sec:organization}
\noindent The remainder of the paper is organized as follows: In section \ref{sec:taxonomy} we present a detailed review of the Video Analytics Techniques that are proposed by different authors to optimize and improve the performance in various applications. Finally, we conclude the paper in Section \ref{sec:conclusions}.

\section{Optimization-based Video Analytics Techniques}
\label{sec:taxonomy}
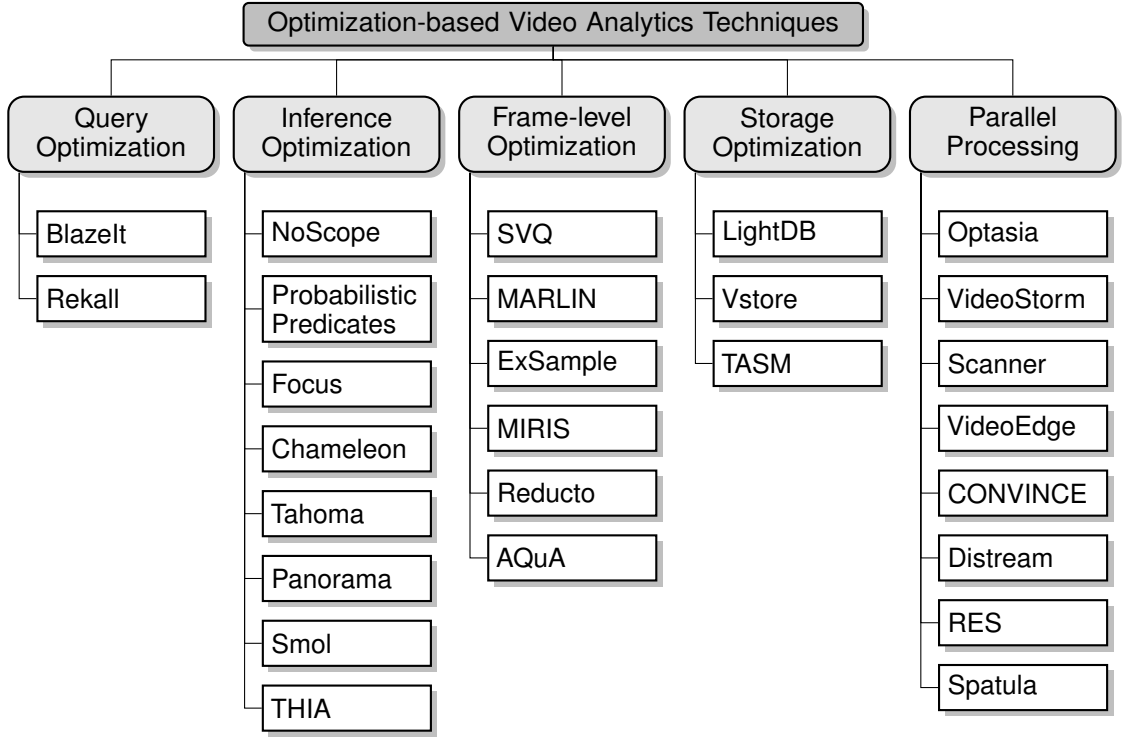
\begin{figure} 
  \centering   
    \begin{forest}
      forked edges,
      for tree={%
        thick,
        anchor=center,
        drop shadow,
        node options={
          draw,
          font=\sffamily
        },
        where level=0{
          parent
        }{
          folder,
          grow'=0,
        },
        where level=1{
          minimum height=1cm,
          child,
          for descendants={%
            grandchild,
            minimum height=0.6cm,
          },
          for children={
            before computing xy={s+=5mm},
          }
        }{},
      }
      [Optimization-based Video Analytics Techniques, xshift=3.6em
        [Query\\ Optimization
            [BlazeIt]
            [Rekall]
        ]
        [Inference\\ Optimization
            [NoScope]
            [Probabilistic Predicates]
            [Focus]
            [Chameleon]
            [Tahoma]
            [Panorama]
            [Smol]
            [THIA]
        ]
        [Frame-level\\ Optimization
            [SVQ]
            [MARLIN]
            [ExSample]
            [MIRIS]
            [Reducto]
            [AQuA]
        ]
        [Storage\\ Optimization
            [LightDB]
            [Vstore]
            [TASM]
        ]
        [Parallel\\ Processing
            [Optasia]
            [VideoStorm]
            [Scanner]
            [VideoEdge]
            [CONVINCE]
            [Distream]
            [RES]
            [Spatula]
        ]
    ]
    \end{forest}
  \caption{Categorization of Video Analytics Techniques based on their optimization strategy}
  \label{fig:taxonomy}
\end{figure}

\noindent Traditional and naive video analytics systems require multiple queries to the learned deep models and suffer from great computational costs. Over the past recent years, researchers have proposed various optimization techniques to improve the performance of video analytics systems. In this section, we review the optimization-based methods that are proposed in video analytics systems to improve performance. We categorize the proposed techniques based on their type of optimization. Note that, techniques in each category are sorted by their year of publication. A diagram of the categorization is shown in Figure \ref{fig:taxonomy}.

\subsection{Query Optimization}
\subsubsection{BlazeIt}
\noindent BlazeIt \cite{kang2018blazeit} is a video analytics system with a declarative query language (called FRAMEQL), an aggregation algorithm, and an algorithm to limit queries. FRAMEQL is mainly used to retrieve the spatio-temporal information of video objects. The aggregation algorithm optimizes the aggregation's efficiency up to 14x compared to existing approximate query processing techniques by using control variates to leverage specialized neural networks. The algorithm that limits queries can have up to 83× speedup compared to recent work in video analytics and random sampling using specialized neural networks. BlazeIt optimizes query execution time by avoiding materialization using proxy models.

\noindent Moreover, authors mention that systems like NoScope \cite{kang2017noscope} and Focus \cite{hsieh2018focus} are inflexible and cannot adapt to the user queries. Authors mention that these systems do not support the extension of specialization and present novel optimization for aggregation and limiting queries, which is supported by BlazeIt.

\subsubsection{Rekall}
\noindent Rekall \cite{fu2019rekall} is a library that exposes a data model and programming model for compositional specifications of video events. The compositional specifications of video events are proposed as a human-in-the-loop approach to detect new interesting events in the video quickly by adapting ideas from multimedia databases and complex event processing over temporal data streams. The experiments show that users who use Rekall can develop queries to retrieve new events in a video given only one hour.

\subsection{Inference Optimization}
\subsubsection{NoScope}
\noindent NoScope \cite{kang2017noscope} is a system for querying videos that accelerates neural network analysis over videos using inference-optimized model search. Given an input video, reference neural network, and the target object, it automatically searches for and trains several models. NoScope takes advantage of two types of models. The first type is specialized models that dispense the generality of standard neural networks in exchange for much faster inference. The second type is difference detectors that identify temporal differences across frames. No Scope combines these two models by performing efficient cost-based optimization to select model architecture and thresholds for each model to maximize throughput subject to a specified accuracy target. NoScope prototype demonstrates speedups of two to three orders of magnitude for binary classification on fixed-angle video streams, with a 1-5\% loss in accuracy.

\subsubsection{Probabilistic Predicates}
\noindent In \cite{lu2018accelerating}, the authors focus on accelerating machine learning inference queries using Probabilistic Predicates, which are executed over the raw input without the need for the predicate column to mirror the original query predicates. The primary use of probabilistic predicates in this work is to filter data blobs that do not satisfy the query predicate. This filtering is parameterized to different target accuracies. The results show that the proposed method has as much as 10× better performance compared to machine learning queries on various large-scale datasets. Moreover, the authors mention that their Probabilistic Predicates strategy differs from NoScope \cite{kang2017noscope} in supporting a wider range of queries and datasets.

\subsubsection{Focus}
\noindent Focus \cite{hsieh2018focus} is a new system that flexibly divides the query processing work between the ingest time and query time. At the ingest time, Focus builds an approximate index of all possible classes of objects in each frame using cheap convolutional neural networks (CNN). It leverages the approximated indices at query time, and it uses an expensive CNNs to compensate for the lower precision. 
Focus reduces the ingest cost on average by 48× (up to 92×) compared to using expensive CNN. Also, Focus makes queries on average 125× (up to 607×) faster than a state-of-the-art video querying system like NoScope \cite{kang2017noscope}.

\subsubsection{Chameleon}
\noindent Chameleon \cite{jiang2018chameleon} is a video analytics system that reduces profiling costs, optimizes resource consumption, and improves inference accuracy of video analytics pipelines. Optimization in Chameleon is done by adapting the configuration of the existing neural network-based video analytics pipelines in real-time. The main idea of adapting the configuration is that the underlying characteristics that affect the best configuration have enough spatio-temporal correlation that amortizes the search cost over time and across multiple video feeds.

\noindent Analysis shows that continuously adapting neural network configurations saves the computation resources by up to 10× and improves accuracy by up to 2× compared to one-time tuning. Moreover, evaluation using the video feeds of five cameras shows that Chameleon can improve the accuracy by 20\% to 50\% with the same amount of resources compared to baseline that picks a single optimal offline configuration, or it can achieve the same accuracy by using only 30\% to 50\% of the resources.

\subsubsection{TAHOMA}
\noindent TAHOMA \cite{anderson2019physical} is a system that accelerates content extraction from extensive visual data to support visual analytics queries. TAHOMA generates and evaluates many cascade classifiers that jointly optimize the Convolutional Neural Network (CNN) architecture and input data representation. The CNN optimization is done by designing and choosing the CNN-based operator that implements a high-quality sensitive relational predicate of images. Mainly, it is done by constructing a large number of cascade classifiers from a large variety of CNN-based classification models.

\noindent Authors show that TAHOMA speeds up classifier cascade through input data transformation by up to 35×. Moreover, it speeds up the ResNet50 image classifier by 98× with no accuracy loss. 

\subsubsection{Panorama}
\noindent Panorama \cite{zhang2019panorama} is the first information system architecture for unbounded vocabulary queries over video. The authors create a new multi-task convolutional neural network (CNN) architecture, which is called PanoramaNet. PanoramaNet supports unbounded vocabularies in a unified and unsupervised manner based on embedding extraction and content-based image retrieval (CBIR). Moreover, they come up with a new self short-circuiting configuration scheme for PanoramaNet to enable practical trade-offs between accuracy and efficiency. 

\noindent The results show that Panorama offers between 2x and 20x higher throughput than competitive accuracy for in-vocabulary queries while also generalizing well with out-of-vocabulary queries. As the vocabulary grows, Panorama saves the users from the difficulty of retraining models post-deployment. The authors also mention the difference between Panorama and NoScope \cite{kang2017noscope}. No Scope uses only two vocabularies, which are “yes” or “no” for a given object type. In Panorama, the number of vocabulary can be infinite, which they call an unbounded vocabulary.

\subsubsection{SMOL}
\noindent SMOL \cite{kang2020jointly} is an engine that focuses on Deep Neural Network (DNN) based visual analytics and optimizes end-to-end query time for DNN inference by optimizing the computational cost of preprocessing and DNN execution. SMOL uses low-resolution data to reduce preprocessing and execution costs, which causes an accuracy reduction. SMOL uses a DNN training procedure that uses data augmentation to recover the accuracy loss caused by using low-resolution data. This work proves that using an accurate large DNN with low-resolution data can be more efficient and higher in accuracy than using a small DNN with high-resolution data.

\noindent The evaluation shows that SMOL can achieve 2.5× throughput improvement at a fixed error level. The comparison of SMOL with BlazeIt shows that SMOL outperforms BlazeIt in all settings.

\subsubsection{THIA}
\noindent THIA \cite{cao2021thia} is a video analytics system that overcomes the limitations of techniques that lower the computational overhead associated with deep learning models. A technique that uses a specialized lightweight model to answer the query has a limitation in providing accurate results for hard-to-detect events. Another technique that filters irrelevant frames using a lightweight model, and then, uses a heavyweight model to process the filtered frames has two limitations: 1) it cannot accelerate the queries that focus on frequently occurring events, and 2) the filter cannot eliminate a significant fraction of video frames.

\noindent THIA works by utilizing three techniques. First, the Early Inference technique constructs a single object detection model with multiple exit points for short-circuiting the inference. This technique offers a set of throughput-accuracy tradeoffs. Second, the Fine-Grained technique uses different exit points to plan and process different video chunks. This technique works simultaneously with the Early Inference technique. Finally, the last technique is the Exit Point Estimation technique, which is a lightweight technique that directly estimates the exit point for a chunk to reduce the optimization overhead of the Fine-Grained Planning technique. Experiments show that THIA outperforms Probabilistic Predicates and BlazeIt systems on a wide range of queries by up to 6.5×. Moreover, it provides accurate query results even on queries that work on hard-to-detect events.

\subsection{Frame-level Optimization}
\subsubsection{SVQ}
\noindent SVQ \cite{xarchakos2019svq} is a system that executes declarative queries on streaming videos. SVQ applies approximate filters to accelerate the query execution that contains specific objects on video frames with spatial relations. These filters use extensible deep neural architectures and easy to deploy and utilize. The results show that the application of filtering techniques with SVQ made the declarative queries on streaming video dramatically increase the frame processing rate. Depending on the query, it speeds up the query processing by at least two orders of magnitude.

\subsubsection{MARLIN}
\noindent MARLIN \cite{apicharttrisorn2019frugal} is a framework to manage and reduce energy consumption for object detection and tracking. It balances between improving tracking accuracy and saving energy by triggering deep neural networks only when needed. The key idea behind MARLIN is to examine only the portions of the frame outside of the currently tracked objects to check if there are any new objects present or recapture objects that significantly change in appearance. Moreover, it ignores the camera motions and effects.

\noindent The results show that MARLIN saves energy by up to 73\%, with at most 7\% accuracy penalty for 75\% of tested videos. Furthermore, for 46.3\% of the cases, MARLIN improves accuracy and reduces the energy consumption compared to systems that continuously use deep neural networks.

\subsubsection{ExSample}
\noindent ExSample \cite{moll2020exsample} is a low-cost framework for object search in an unindexed video. ExSample processes search queries quickly by adapting the amount and location of sampled frames to the specified data and query that is being processed. Frame sampling is done to find the most distinctive objects in the shortest amount of time. ExSample iteratively processes batches of frames. Furthermore, tuning the sampling process is based on whether new objects are found in the previous frames sampled on each iteration.

\noindent The evaluation of ExSample shows that it can reduce both the number of sampled frames and the execution time needed to achieve a particular recall. The authors mention that ExSample works better in time and number of frames compared to surrogate models, such as BlazeIt. BlazeIt selects frames for applying the object detector based on surrogate scores. However, ExSample works with an adaptive sampling technique that eliminates the need for a surrogate model.

\subsubsection{MIRIS}
\noindent MIRIS \cite{bastani2020miris} is a video query processor that integrates query processing and object tracking to select a variable video sampling frame rate that minimizes object detector workload while maintaining accurate query outputs. The Evaluation of MIRIS shows that it speeds up the object tracker execution by 9×, and it reduces the number of video frames that must be processed.

\noindent The authors compare MIRIS with NoScope \cite{kang2017noscope}, probabilistic predicates \cite{lu2018accelerating}, BlazeIt \cite{kang2018blazeit}, and SVQ \cite{xarchakos2019svq}. These engines train lightweight specialized machine learning models to approximate the result of a predicate over individual video frames or sequence of frames. The authors also mention two issues of these engines. First, the object instance that satisfies the query predicate, most of the time, appears in almost all frames, which will make it difficult for these engines to offer a substantial speedup. Moreover, object tracking queries inherently involve a video sequence, which leads to expensive object detectors on almost every video frame.

\subsubsection{Reducto}
\noindent Reducto \cite{li2020reducto} is a video analytics system that performs on-camera frame filtering while supporting resource efficient real-time querying for video analytics. Reducto works by dynamically adapting filtering decisions according to the time-varying correlation between video feature type, filtering threshold, query accuracy, and video content. Reducto selects low-level video features by determining the best feature for each query class. It also chooses a filtering threshold using a lightweight machine learning technique to predict the chosen feature threshold while maintaining query accuracy. Evaluations show that Reducto achieves significant filtering benefits while meeting the desired accuracy. 

\subsubsection{AQuA}
\noindent AQuA \cite{paul2021aqua} is a deep learning model that protects the accuracy of the application against poor quality frames by scoring the frames distortion level and assigning an analytical quality score according to whether the frame is good or bad for further analysis. AQuA can detect, score, or discard the distorted frames either after capture at the edge or after video compression and transmission. AQuA works by using classifier opinion scores to evaluate an analytical quality of a frame. The authors mention that AQuA is the first system to improve the real-time video analysis pipeline by considering a classifier assessment of image quality.

\noindent The evaluation of AQuA shows that it reduces high-confidence errors for analytics applications by up to 17\% when filtering poor quality and distorted frames at the edge. Moreover, it reduces computation time and average bandwidth usage by up to 25\%.

\subsection{Storage Optimization}
\subsubsection{LightDB}
\noindent LightDB \cite{haynes2018lightdb} is a database management system that efficiently manages virtual, augmented, and mixed reality (VAMR) video content. In LightDB, VAMR is treated as a six-dimensional light field. Besides, LightDB supports a rich set of operations over light fields, and it automatically transforms declarative queries into executable physical plans. Experimental results show that LightDB offers up to 4× throughput improvements compared to prior work. Moreover, it has easily expressible queries, and it improves the performance of queries up to 500× compared to other video processing frameworks. Authors also mention that LightDB can process up to 8× more frames per second than the Scanner method.

\subsubsection{VStore}
\noindent VStore \cite{xu2019vstore} is a data store that manages video ingestion, storage, retrieval, and video resource usage. VStore supports fast and efficient video analysis over large videos and it controls the video format along the data path. VStore works with an idea called backward derivation of configuration that works by passing the desired video quality and quantity back to retrieval, storage, and ingestion stages. Results show that VStore runs queries as fast as 362x of video runtime. Furthermore, authors mention that VStore is the first holistic system that manages the full video lifecycle for retrospective analytics.

\subsubsection{TASM}
\noindent TASM \cite{daum2020tasm} is a storage manager for video data that improves video query performance. TASM speeds up queries that retrieve objects in a video with low storage overhead and good video quality by splitting the video frames into independent tiles and optimizes the video file layout based on its content and the query workload. TASM designs tile layouts to improve video query performance by including information about the video content with the observation of the objects targeted by queries. Evaluations show that layouts picked by TASM accelerate individual queries by an average of 51\% and up to 94\% while maintaining good quality. Moreover, TASM can automatically adjust layouts over a small number of queries to improve query performance even for the unknown query workloads. 

\subsection{Parallel Processing}
\subsubsection{Optasia}
\noindent Optasia \cite{lu2016optasia} is a system that combines the most recent techniques from vision and data-parallel computing communities mainly for various surveillance applications. It provides a SQL-like declarative language. Moreover, it works with a cost-based query optimizer (QO) to connect and take benefits from both end-user queries and low-level vision modules. QO has several advantages. For instance, it outputs good parallel execution plans, it scales appropriately as the data size increases, and it helps in reducing the duplication of overall work by structuring the work of each query. Authors show that Optasia improves the accuracy and performance several times more than prior works on surveillance videos.

\subsubsection{VideoStorm}
\noindent VideoStorm \cite{zhang2017live} is a system for video analytics that scales to processing thousands of video queries on live video streams over large clusters. In this work, VideoStorm is deployed on an Azure cluster of 101 machines. Initially, users submit video queries that contain many arbitrary vision processors. Afterwards, VideoStorm generates query resource-quality profiles for different query knobs configurations. Simultaneously, it uses its scheduler to improve performance by maximizing the quality and minimizing the lag tolerance on video queries in allocating resources.

\noindent Results show that generating query profiles uses 3.5× fewer CPU resources compared to a basic greedy search. The scheduler performs fair scheduling as much as 80\% in quality of real-world queries and 7× better in terms of lag.

\subsubsection{Scanner}
\noindent Scanner \cite{poms2018scanner} is a system for efficient video analysis at scale. It supports two important aspects of video analysis. First, storing and accessing pixel data from several large videos by organizing video collections as tables in a data store, whose implementation is optimized for compressed video. Second, executing expensive pixel-level operations in parallel by organizing pixel-analysis tasks as dataflow graphs that operate on sequences of frames sampled from tables. The Evaluation of Scanner shows that video analysis tasks that require days of processing can be done efficiently in hours to minutes. 

\subsubsection{VideoEdge}
\noindent VideoEdge \cite{hung2018videoedge} is a system that identifies the best trade-off between multiple resources and accuracy. VideoEdge narrows search space by identifying the most promising options in the “Pareto band” and searches only within the band. Moreover, as video analysis queries have multiple implementations, VideoEdge decides the implementation and the knobs of queries, places the queries across the hierarchy, and merges queries with common processing. It balances the resource benefits and accuracy drawback of merging queries. Results show that VideoEdge improves accuracy by 25.4× compared to the fair distribution of resources and 5.4x compared to a recent video query planning solution.

\subsubsection{CONVINCE}
\noindent CONVINCE \cite{pasandi2020convince} is a cross-camera video analytics system that enables a collaborative video analytics pipeline among network-connected cameras. CONVINCE works by leveraging spatio-temporal correlations and knowledge sharing while preserving privacy. Spatio-temporal correlations are done by discarding redundant frames to reduce the cost of bandwidth and processing. Knowledge sharing is done to improve the accuracy of vision models. Results show that CONVINCE achieves 91\% accuracy of object identification by transmitting only about 25\% of all recorded frames.

\subsubsection{Distream}
\noindent Distream \cite{zeng2020distream} is a framework for distributed live video analytics based on the smart camera-edge cluster architecture. The main advantage of Distream is its ability to adapt the real-world workload dynamics to achieve low latency, high throughput, and scalable deep learning-based live video analytics. Distream is designed to balance the workloads between smart cameras, partition the workload between the cameras and the cluster, and adapt the workload dynamics.

\noindent Authors evaluate Distream using 24 cameras and a 4-GPU edge cluster on 500 hours of a distributed video stream from two real-world video datasets. The evaluation results show that Distream outperforms the existing methods in throughput, latency, and service level objective (SLO) miss rate.

\subsubsection{RES}
\noindent RES \cite{ali2020res} is an edge-enhanced stream analysis system that complements a cloud-based platform for video stream analytics. RES works in two phases, while each phase consists of three stages. The first phase is the filtration phase, which reduces data by detecting and filtering low-value objects using user configuration rules. The second phase is the identification phase, which applies deep learning to objects of interest for further analysis. The three stages for both phases are basic, filter, and machine learning, that analyze and partition the video analysis pipeline. RES distributes the processing stages over available resources in the cloud to meet the user’s real-time quality of service (QoS) requirements. Experiment on a 10K datastream shows that RES reduces the time by 49\% and saves 99\% of bandwidth compared to a centralized cloud-based analytics approach.

\subsubsection{Spatula}
\noindent Spatula \cite{jain2020spatula} is a cross-camera analytics system that reduces the network and computation costs by leveraging the spatio-temporal cross-camera correlation. The main idea behind Spatula is to use the spatio-temporal correlation to limit the analyzed data. Limiting data is done by streaming and running the cross-camera inference on only the set of cameras and frames that contain the queried object and not the total number of deployed cameras, which in turn, reduces the cross-camera analytics cost. Evaluation on an 8-camera dataset shows that Spatula reduces the computation workload by 8.3× and improves the inference precision by 39\%. Moreover, on two datasets with hundreds of cameras, it reduces the computation workload by 23× to 86×.

\section{Conclusion}
\label{sec:conclusions}
\noindent Video Analytics is an essential topic which deals with detecting objects in video, predicting the next steps of the moving objects, and understanding the behavior of objects by tracking them. Video Analytics is used in many fields such as self-driving cars, autonomous drones, and safety and security applications. In this survey, we reviewed the most recent Video Analytics techniques which focus in optimizing the performance. In our review, we categorized the techniques based on the type of optimization that they utilize. Specifically, this survey covers the definitions of each technique, how it works, how it improves performance, and the improved results.

\bibliographystyle{plain}  

\begin{thebibliography}{10}

    \bibitem{ali2020res}
    Muhammad Ali, Ashiq Anjum, Omer Rana, Ali~Reza Zamani, Daniel Balouek-Thomert,
      and Manish Parashar.
    \newblock Res: Real-time video stream analytics using edge enhanced clouds.
    \newblock {\em IEEE Transactions on Cloud Computing}, 2020.

    \bibitem{anderson2019physical}
    Michael~R Anderson, Michael Cafarella, German Ros, and Thomas~F Wenisch.
    \newblock Physical representation-based predicate optimization for a visual
      analytics database.
    \newblock In {\em 2019 IEEE 35th International Conference on Data Engineering
      (ICDE)}, pages 1466--1477. IEEE, 2019.

    \bibitem{apicharttrisorn2019frugal}
    Kittipat Apicharttrisorn, Xukan Ran, Jiasi Chen, Srikanth~V Krishnamurthy, and
      Amit~K Roy-Chowdhury.
    \newblock Frugal following: Power thrifty object detection and tracking for
      mobile augmented reality.
    \newblock In {\em Proceedings of the 17th Conference on Embedded Networked
      Sensor Systems}, pages 96--109, 2019.

    \bibitem{bastani2020miris}
    Favyen Bastani, Songtao He, Arjun Balasingam, Karthik Gopalakrishnan, Mohammad
      Alizadeh, Hari Balakrishnan, Michael Cafarella, Tim Kraska, and Sam Madden.
    \newblock Miris: Fast object track queries in video.
    \newblock In {\em Proceedings of the 2020 ACM SIGMOD International Conference
      on Management of Data}, pages 1907--1921, 2020.

    \bibitem{bastani2020vaas}
    Favyen Bastani, Oscar Moll, and Sam Madden.
    \newblock Vaas: video analytics at scale.
    \newblock {\em Proceedings of the VLDB Endowment}, 13(12):2877--2880, 2020.

    \bibitem{cao2021thia}
    Jiashen Cao, Ramyad Hadidi, Joy Arulraj, and Hyesoon Kim.
    \newblock Thia: Accelerating video analytics using early inference and
      fine-grained query planning.
    \newblock {\em arXiv preprint arXiv:2102.08481}, 2021.

    \bibitem{cuevas2020techniques}
    Carlos Cuevas, Daniel Quilon, and Narciso Garcia.
    \newblock Techniques and applications for soccer video analysis: A survey.
    \newblock {\em Multimedia Tools and Applications}, 79(39):29685--29721, 2020.

    \bibitem{daum2020tasm}
    Maureen Daum, Brandon Haynes, Dong He, Amrita Mazumdar, Magdalena Balazinska,
      and Alvin Cheung.
    \newblock Tasm: A tile-based storage manager for video analytics.
    \newblock {\em arXiv preprint arXiv:2006.02958}, 2020.

    \bibitem{fu2019rekall}
    Daniel~Y Fu, Will Crichton, James Hong, Xinwei Yao, Haotian Zhang, Anh Truong,
      Avanika Narayan, Maneesh Agrawala, Christopher R{\'e}, and Kayvon Fatahalian.
    \newblock Rekall: Specifying video events using compositions of spatiotemporal
      labels.
    \newblock {\em arXiv preprint arXiv:1910.02993}, 2019.

    \bibitem{haynes2018lightdb}
    Brandon Haynes, Amrita Mazumdar, Magdalena Balazinska, Luis Ceze, and Alvin
      Cheung.
    \newblock Lightdb: A dbms for virtual reality video.
    \newblock {\em Proceedings of the VLDB Endowment}, 11(10), 2018.

    \bibitem{hsieh2018focus}
    Kevin Hsieh, Ganesh Ananthanarayanan, Peter Bodik, Shivaram Venkataraman,
      Paramvir Bahl, Matthai Philipose, Phillip~B Gibbons, and Onur Mutlu.
    \newblock Focus: Querying large video datasets with low latency and low cost.
    \newblock In {\em 13th $\{$USENIX$\}$ Symposium on Operating Systems Design and
      Implementation ($\{$OSDI$\}$ 18)}, pages 269--286, 2018.

    \bibitem{hung2018videoedge}
    Chien-Chun Hung, Ganesh Ananthanarayanan, Peter Bodik, Leana Golubchik, Minlan
      Yu, Paramvir Bahl, and Matthai Philipose.
    \newblock Videoedge: Processing camera streams using hierarchical clusters.
    \newblock In {\em 2018 IEEE/ACM Symposium on Edge Computing (SEC)}, pages
      115--131. IEEE, 2018.

    \bibitem{jain2020spatula}
    Samvit Jain, Xun Zhang, Yuhao Zhou, Ganesh Ananthanarayanan, Junchen Jiang,
      Yuanchao Shu, Paramvir Bahl, and Joseph Gonzalez.
    \newblock Spatula: Efficient cross-camera video analytics on large camera
      networks.
    \newblock In {\em 2020 IEEE/ACM Symposium on Edge Computing (SEC)}, pages
      110--124. IEEE, 2020.

    \bibitem{jiang2018chameleon}
    Junchen Jiang, Ganesh Ananthanarayanan, Peter Bodik, Siddhartha Sen, and Ion
      Stoica.
    \newblock Chameleon: scalable adaptation of video analytics.
    \newblock In {\em Proceedings of the 2018 Conference of the ACM Special
      Interest Group on Data Communication}, pages 253--266, 2018.

    \bibitem{kang2018blazeit}
    Daniel Kang, Peter Bailis, and Matei Zaharia.
    \newblock Blazeit: optimizing declarative aggregation and limit queries for
      neural network-based video analytics.
    \newblock {\em arXiv preprint arXiv:1805.01046}, 2018.

    \bibitem{kang2017noscope}
    Daniel Kang, John Emmons, Firas Abuzaid, Peter Bailis, and Matei Zaharia.
    \newblock Noscope: optimizing neural network queries over video at scale.
    \newblock {\em arXiv preprint arXiv:1703.02529}, 2017.

    \bibitem{kang2020jointly}
    Daniel Kang, Ankit Mathur, Teja Veeramacheneni, Peter Bailis, and Matei
      Zaharia.
    \newblock Jointly optimizing preprocessing and inference for dnn-based visual
      analytics.
    \newblock {\em arXiv preprint arXiv:2007.13005}, 2020.

    \bibitem{kong2018human}
    Yu~Kong and Yun Fu.
    \newblock Human action recognition and prediction: A survey.
    \newblock {\em arXiv preprint arXiv:1806.11230}, 2018.

    \bibitem{kulkarni2020survey}
    Smita Kulkarni, Sangeeta Jadhav, and Debashis Adhikari.
    \newblock A survey on human group activity recognition by analysing person
      action from video sequences using machine learning techniques.
    \newblock {\em Optimization in Machine Learning and Applications}, pages
      141--153, 2020.

    \bibitem{li2020reducto}
    Yuanqi Li, Arthi Padmanabhan, Pengzhan Zhao, Yufei Wang, Guoqing~Harry Xu, and
      Ravi Netravali.
    \newblock Reducto: On-camera filtering for resource-efficient real-time video
      analytics.
    \newblock In {\em Proceedings of the Annual conference of the ACM Special
      Interest Group on Data Communication on the applications, technologies,
      architectures, and protocols for computer communication}, pages 359--376,
      2020.

    \bibitem{lu2016optasia}
    Yao Lu, Aakanksha Chowdhery, and Srikanth Kandula.
    \newblock Optasia: A relational platform for efficient large-scale video
      analytics.
    \newblock In {\em Proceedings of the Seventh ACM Symposium on Cloud Computing},
      pages 57--70, 2016.

    \bibitem{lu2018accelerating}
    Yao Lu, Aakanksha Chowdhery, Srikanth Kandula, and Surajit Chaudhuri.
    \newblock Accelerating machine learning inference with probabilistic
      predicates.
    \newblock In {\em Proceedings of the 2018 International Conference on
      Management of Data}, pages 1493--1508, 2018.

    \bibitem{madigan2019understanding}
    Ruth Madigan, Sina Nordhoff, Charles Fox, Roja~Ezzati Amini, Tyron Louw, Marc
      Wilbrink, Anna Schieben, and Natasha Merat.
    \newblock Understanding interactions between automated road transport systems
      and other road users: a video analysis.
    \newblock {\em Transportation research part F: traffic psychology and
      behaviour}, 66:196--213, 2019.

    \bibitem{moll2020exsample}
    Oscar Moll, Favyen Bastani, Sam Madden, Mike Stonebraker, Vijay Gadepally, and
      Tim Kraska.
    \newblock Exsample: Efficient searches on video repositories through adaptive
      sampling.
    \newblock {\em arXiv preprint arXiv:2005.09141}, 2020.

    \bibitem{okita2020ai}
    Hideki Okita, Tomokazu Murakami, Tatsuya Okubo, Toshiaki Tarui, Yasuhiro
      Fukuda, and PE~Jp.
    \newblock Ai-based video analysis solution for creating safe and secure
      society.
    \newblock {\em Hitachi Review}, 69:687--693, 2020.

    \bibitem{olatunji2019video}
    Iyiola~E Olatunji and Chun-Hung Cheng.
    \newblock Video analytics for visual surveillance and applications: An overview
      and survey.
    \newblock {\em Machine Learning Paradigms}, pages 475--515, 2019.

    \bibitem{pasandi2020convince}
    Hannaneh~Barahouei Pasandi and Tamer Nadeem.
    \newblock Convince: Collaborative cross-camera video analytics at the edge.
    \newblock In {\em 2020 IEEE International Conference on Pervasive Computing and
      Communications Workshops (PerCom Workshops)}, pages 1--5. IEEE, 2020.

    \bibitem{paul2021aqua}
    Sibendu Paul, Utsav Drolia, Y~Charlie Hu, and Srimat~T Chakradhar.
    \newblock Aqua: Analytical quality assessment for optimizing video analytics
      systems.
    \newblock {\em arXiv preprint arXiv:2101.09752}, 2021.

    \bibitem{poddar2020visor}
    Rishabh Poddar, Ganesh Ananthanarayanan, Srinath Setty, Stavros Volos, and
      Raluca~Ada Popa.
    \newblock Visor: Privacy-preserving video analytics as a cloud service.
    \newblock In {\em 29th $\{$USENIX$\}$ Security Symposium ($\{$USENIX$\}$
      Security 20)}, pages 1039--1056, 2020.

    \bibitem{poms2018scanner}
    Alex Poms, Will Crichton, Pat Hanrahan, and Kayvon Fatahalian.
    \newblock Scanner: Efficient video analysis at scale.
    \newblock {\em ACM Transactions on Graphics (TOG)}, 37(4):1--13, 2018.

    \bibitem{premkumar2021video}
    Sree Premkumar, Vimal Premkumar, and Rakesh Dhakshinamurthy.
    \newblock Video analytics on iot devices.
    \newblock {\em arXiv preprint arXiv:2102.07455}, 2021.

    \bibitem{ran2018deepdecision}
    Xukan Ran, Haolianz Chen, Xiaodan Zhu, Zhenming Liu, and Jiasi Chen.
    \newblock Deepdecision: A mobile deep learning framework for edge video
      analytics.
    \newblock In {\em IEEE INFOCOM 2018-IEEE Conference on Computer
      Communications}, pages 1421--1429. IEEE, 2018.

    \bibitem{shih2017survey}
    Huang-Chia Shih.
    \newblock A survey of content-aware video analysis for sports.
    \newblock {\em IEEE Transactions on Circuits and Systems for Video Technology},
      28(5):1212--1231, 2017.

    \bibitem{suprem2020odin}
    Abhijit Suprem, Joy Arulraj, Calton Pu, and Joao Ferreira.
    \newblock Odin: automated drift detection and recovery in video analytics.
    \newblock {\em arXiv preprint arXiv:2009.05440}, 2020.

    \bibitem{wang2003video}
    Jessica~JunLin Wang and Sameer Singh.
    \newblock Video analysis of human dynamics—a survey.
    \newblock {\em Real-time imaging}, 9(5):321--346, 2003.

    \bibitem{xarchakos2019svq}
    Ioannis Xarchakos and Nick Koudas.
    \newblock Svq: Streaming video queries.
    \newblock In {\em Proceedings of the 2019 International Conference on
      Management of Data}, pages 2013--2016, 2019.

    \bibitem{xu2019vstore}
    Tiantu Xu, Luis~Materon Botelho, and Felix~Xiaozhu Lin.
    \newblock Vstore: A data store for analytics on large videos.
    \newblock In {\em Proceedings of the Fourteenth EuroSys Conference 2019}, pages
      1--17, 2019.

    \bibitem{zeng2020distream}
    Xiao Zeng, Biyi Fang, Haichen Shen, and Mi~Zhang.
    \newblock Distream: scaling live video analytics with workload-adaptive
      distributed edge intelligence.
    \newblock In {\em Proceedings of the 18th Conference on Embedded Networked
      Sensor Systems}, pages 409--421, 2020.

    \bibitem{zhang2017live}
    Haoyu Zhang, Ganesh Ananthanarayanan, Peter Bodik, Matthai Philipose, Paramvir
      Bahl, and Michael~J Freedman.
    \newblock Live video analytics at scale with approximation and delay-tolerance.
    \newblock In {\em 14th $\{$USENIX$\}$ Symposium on Networked Systems Design and
      Implementation ($\{$NSDI$\}$ 17)}, pages 377--392, 2017.

    \bibitem{zhang2019edge}
    Qingyang Zhang, Hui Sun, Xiaopei Wu, and Hong Zhong.
    \newblock Edge video analytics for public safety: A review.
    \newblock {\em Proceedings of the IEEE}, 107(8):1675--1696, 2019.

    \bibitem{zhang2018physics}
    Xuguang Zhang, Qinan Yu, and Hui Yu.
    \newblock Physics inspired methods for crowd video surveillance and analysis: a
      survey.
    \newblock {\em IEEE Access}, 6:66816--66830, 2018.

    \bibitem{zhang2019panorama}
    Yuhao Zhang and Arun Kumar.
    \newblock Panorama: a data system for unbounded vocabulary querying over video.
    \newblock {\em Proceedings of the VLDB Endowment}, 13(4):477--491, 2019.

    \bibitem{zhang2020machine}
    Yun Zhang, Sam Kwong, and Shiqi Wang.
    \newblock Machine learning based video coding optimizations: A survey.
    \newblock {\em Information Sciences}, 506:395--423, 2020.

\end{thebibliography}

\end{document}